\documentclass[nohyperref]{article}

\usepackage{microtype}
\usepackage{graphicx}
\usepackage{subfigure}
\usepackage{booktabs} 
\usepackage{enumitem}

\usepackage{hyperref}

 \usepackage[accepted]{icml2022}

\usepackage{amsmath}
\usepackage{amssymb}
\usepackage{mathtools}
\usepackage{amsthm}

\usepackage[capitalize,noabbrev]{cleveref}

\theoremstyle{plain}
\newtheorem{theorem}{Theorem}[section]

\theoremstyle{definition}
\newtheorem{definition}[theorem]{Definition}

\theoremstyle{remark}

\usepackage[textsize=tiny]{todonotes}

\icmltitlerunning{Sheaf Neural Networks with Connection Laplacians}

\usepackage{import}
\newcommand{\hide}[1]{} 

\newcommand{\R}{\mathbb{R}}

\newcommand{\Fs}{\mathcal{F}}

\newcommand{\Ms}{\mathcal{M}}
\newcommand{\Ns}{\mathcal{N}}

\newcommand{\Wb}{\textbf{W}}
\newcommand{\xb}{\textbf{x}}
\newcommand{\Xb}{\textbf{X}}
\newcommand{\Ib}{\textbf{I}}

\newcommand{\Ab}{\textbf{A}}
\newcommand{\Bb}{\textbf{B}}

\newcommand{\Db}{\textbf{D}}
\newcommand{\Ub}{\textbf{U}}
\newcommand{\Hb}{\textbf{H}}
\renewcommand{\Bb}{\textbf{B}}
\newcommand{\Vb}{\textbf{V}}
\newcommand{\Ob}{\textbf{O}}

\newcommand{\incident}{\trianglelefteq}

\newenvironment{talign}{\align}{\endalign}
\newenvironment{talign*}{\csname align*\endcsname}{\endalign}

\begin{document}

\twocolumn[
\icmltitle{Sheaf Neural Networks with Connection Laplacians}



\icmlsetsymbol{equal}{*}

\begin{icmlauthorlist}
\icmlauthor{Federico Barbero}{camb}
\icmlauthor{Cristian Bodnar}{camb}
\icmlauthor{Haitz S\'aez de Oc\'ariz Borde}{camb} \\
\icmlauthor{Michael Bronstein}{michael}
\icmlauthor{Petar Veli\v{c}kovi\'c}{petar}
\icmlauthor{Pietro Li\`o}{camb}
\end{icmlauthorlist}

\icmlaffiliation{camb}{University of Cambridge}
\icmlaffiliation{michael}{University of Oxford \& Twitter}
\icmlaffiliation{petar}{DeepMind}

\icmlcorrespondingauthor{Federico Barbero}{fb548@cam.ac.uk}

\icmlkeywords{Machine Learning, ICML}

\vskip 0.3in
]



\printAffiliationsAndNotice{}
\begin{abstract}
A Sheaf Neural Network (SNN) is a type of Graph Neural Network (GNN) that operates on a sheaf, an object that equips a graph with vector spaces over its nodes and edges and linear maps between these spaces. SNNs have been shown to have useful theoretical properties that help tackle issues arising from heterophily and over-smoothing. One complication intrinsic to these models is finding a good sheaf for the task to be solved. Previous works proposed two diametrically opposed approaches: manually constructing the sheaf based on domain knowledge and learning the sheaf end-to-end using gradient-based methods. However, domain knowledge is often insufficient, while learning a sheaf could lead to overfitting and significant computational overhead. In this work, we propose a novel way of computing sheaves drawing inspiration from Riemannian geometry: we leverage the manifold assumption to compute manifold-and-graph-aware orthogonal maps, which optimally align the tangent spaces of neighbouring data points. We show that this approach achieves promising results with less computational overhead when compared to previous SNN models. Overall, this work provides an interesting connection between algebraic topology and differential geometry, and we hope that it will spark future research in this direction.
\end{abstract}

\section{Introduction}
Graph Neural Networks (GNNs) \cite{scarselli2008graph} have shown encouraging results in a wide range of applications, ranging from drug design \cite{Stokes2020ADL} to guiding discoveries in pure mathematics \cite{davies2021advancing}. One advantage over traditional neural networks is that they can leverage the extra structure in graph data, such as edge connections. 

GNNs, however, do not come without issues. Traditional GNN models, such as Graph Convolutional Networks~(GCNs) \cite{kipf2016semi} have been shown to work poorly on heterophilic data. In fact, GCNs use homophily as an inductive bias by design, that is, they assume that connected nodes will likely belong to the same class and have similar feature vectors, which is not true in many real-world applications~\citep{zhu2020beyond}. Moreover, GNNs also suffer from over-smoothing \cite{oono2019graph}, which prevents these models from improving, and may actually even worsen their performance when stacking several layers. These two problems are, from a geometric point of view, intimately connected \cite{chen2020measuring, bodnar2022neural}.

\begin{figure}[t]
\vskip 0.2in
\begin{center}
\centerline{\includegraphics[width=0.8\columnwidth]{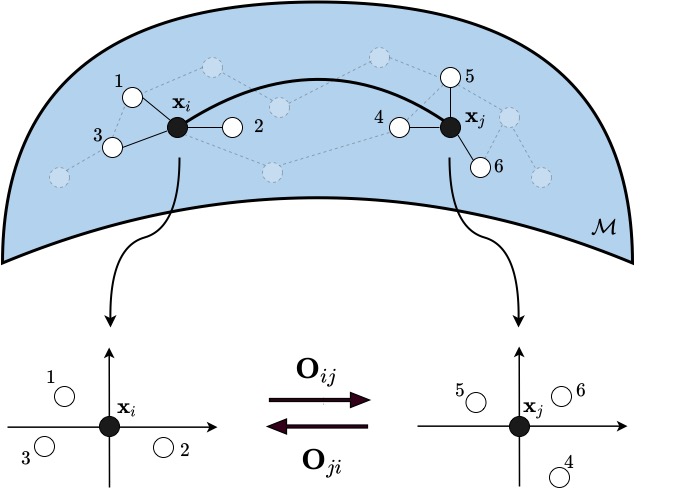}}
\caption{The orthonormal bases of $T_{\mathbf{x}_i}\Ms$ and $T_{\mathbf{x}_j}\Ms$ are determined by local PCA using nodes in the 1-hop neighbourhood of $\mathbf{x}_i$ and $\mathbf{x}_j$ respectively. The orthogonal mapping $O_{ij}$ is a map from $T_{\mathbf{x}_i}\Ms$ to $T_{\mathbf{x}_j}\Ms$ which optimally aligns their bases.}
\label{fig:local-pca}
\end{center}
\vspace{-25pt} 
\end{figure}

\citet{bodnar2022neural} showed that when the underlying ``geometry'' of the graph is too simple, the issues discussed above arise. More precisely, they analysed the geometry of the graph through cellular sheaf theory \cite{curry2014sheaves, hansen2020laplacians, hansen2019toward}, a subfield of algebraic topology \cite{hatcher2005algebraic}. A cellular sheaf associates a vector space to each node and edge of a graph, and linear maps between these spaces. A GNN which operates over a cellular sheaf is known as a Sheaf Neural Network (SNN) \cite{hansen2020sheaf, bodnar2022neural}. 

SNNs work by computing a sheaf Laplacian, which recovers the well-known graph Laplacian when the underlying sheaf is trivial, that is, when vector spaces are 1-dimensional and we apply identity maps between them. \citet{hansen2019toward} have first shown the utility of SNNs in a toy experimental setting, where they used a manually-constructed sheaf Laplacian based on full knowledge of the data generation process. \citet{bodnar2022neural} proposed to \emph{learn} this sheaf Laplacian from data using stochastic gradient descent, making these types of models applicable to any graph dataset. However, this can also lead to computational complexity problems, overfitting and optimisation issues. 

This work proposes a novel technique that aims to precompute a sheaf Laplacian from data in a deterministic manner, removing the need to learn it with gradient-based approaches. We do this through the lens of differential geometry, by assuming that the data is sampled from a low-dimensional manifold and optimally aligning the neighbouring tangent spaces via orthogonal transformations ( see Figure~\ref{fig:local-pca}). This idea was first introduced as groundwork for vector diffusion maps by \citet{singer2012vector}. However, it only assumed a point-cloud structure. Instead, one of our contributions involves the computation of these optimal alignments over a graph structure. We find that our proposed technique performs well, while reducing the computational overhead involved in learning the sheaf.

In Section \ref{sec:background}, we present a brief overview of cellular sheaf theory and neural sheaf diffusion \cite{bodnar2022neural}. Next, in Section \ref{sec:new-technique}, we give details of our new procedure used to pre-compute the sheaf Laplacian before the model-training phase, which we refer to as Neural Sheaf Diffusion with Connection Laplacians (Conn-NSD). We then, in Section \ref{sec:eval}, evaluate this technique on various datasets with varying homophily levels. We believe that this work is a promising attempt at connecting ideas from algebraic topology and differential geometry with machine learning, and hope that it will spark further research at their intersection. 

\section{Background}
\label{sec:background}

We briefly overview the necessary background, starting with GNNs and cellular sheaf theory and concluding with neural sheaf diffusion. The curious reader may refer to \citet{curry2014sheaves, hansen2020laplacians, hansen2019toward} for a more in-depth insight into cellular sheaf theory, and to \citet{bodnar2022neural} for the full theoretical results of neural sheaf diffusion.

\subsection{Graph Neural Networks}
GNNs are a family of neural network architectures that generalise neural networks to arbitrarily structured graphs. A graph $G=(V,E)$ is a tuple consisting of a set of nodes $V$ and a set of edges $E$. We can represent each node in the graph with a $d$-dimensional feature vector $\xb_v$ and group all the $n = |V|$ feature vectors into a $n \times d$ matrix $\Xb$. We represent the set of edges $E$ with an adjacency matrix $\Ab$. A GNN layer then takes these two matrices as input to produce a new set of (latent) feature vectors for each node:
\begin{equation}
\label{eq:gnn}
    \Hb^{(l)} = f\left(\Hb^{(l-1)}, \Ab\right).
\end{equation}
In the case of a multi-layer GNN, the first layer $l=1$, takes as input $\Hb^{(0)} = \Xb$, whereas subsequent layers, $l$, take as input $\Hb^{(l-1)}$, the latent features produced by the GNN layer immediately before it. There are numerous architectures which take this form, with one of the most popular being the Graph Convolutional Network (GCN) \cite{kipf2016semi} which implements Equation (\ref{eq:gnn}) the following way:
\begin{equation}
    \label{eq:gcn}
    \Hb^{(l)} = \sigma \left(\hat{\Db}^{-\frac{1}{2}}\hat{\Ab}\hat{\Db}^{-\frac{1}{2}}\Hb^{(l-1)}\Wb^{(l)}\right),
\end{equation}
where $\sigma$ is a non-linear activation function (e.g. ReLU), $\hat{\Ab} = \Ab + \Ib$, $\hat{\Db}$ is the diagonal node degree matrix of $\hat{\Ab}$ and $\Wb^{(l)}$ is a weight matrix. This update propagation is local (due to the adjacency matrix), meaning that each latent feature vector is updated as a function of its local neighbourhood, weighted by a weight matrix and then symmetrically normalised. This kind of model has proven to be extremely powerful in a myriad of tasks. The weight matrix $\Wb^{(l)}$ at each layer is learnt from the data through back-propagation, by minimising some loss function (e.g. cross-entropy loss).
\subsection{Cellular Sheaf Theory}

\begin{definition}
A cellular sheaf $(G, \Fs)$ on an \emph{undirected graph} $G=(V,E)$ consists of:
\begin{itemize}[leftmargin=5mm, topsep=0pt,itemsep=-0.4ex]
    \item A vector space $\Fs(v)$ for each $v \in V$,
    \item A vector space $\Fs(e)$ for each $e \in E$,
    \item A linear map $\Fs_{v \incident e} : \Fs(v) \to \Fs(e)$ for each incident node-edge pair $v \incident e$.
\end{itemize}
\end{definition}

The vector spaces of the node and edges are called \emph{stalks}, while the linear maps are called \emph{restriction maps}. It is then natural to group the various spaces. The space which is formed by the node stalks is called the space of $0$-cochains, while the space formed by edge stalks is called the space of $1$-cochains.


\begin{definition}
Given a sheaf $(G, \Fs)$, we define the space of $0$-cochains $C^0(G, \Fs)$ as the direct sum over the vertex stalks $C^0(G, \Fs) := \bigoplus_{v \in V} \Fs(v)$. Similarly, the space of $1$-cochains $C^1(G, \Fs)$ as the direct sum over the edge stalks $C^1(G, \Fs)  := \bigoplus_{e \in E} \Fs(e)$. 
\end{definition}

Defining the spaces $C^0(G, \Fs)$ and $C^1(G, \Fs)$ allows us to construct a linear \emph{co-boundary map} $\delta : C^0(G, \Fs) \to C^1(G, \Fs)$. From an opinion dynamics perspective \cite{hansen2021opinion},  the node stalks may be thought of as the private space of opinions and the edge stalks as the space in which these opinions are shared in a public discourse space. The co-boundary map $\delta$ then measures the disagreement between all the nodes.

\begin{definition}
Given some arbitrary orientation for each edge $e = u \to v \in E$, we define the co-boundary map $\delta : C^0(G, \Fs) \to C^1(G, \Fs)$ as $\delta(\xb)_e = \Fs_{v \incident e}\xb_v - \Fs_{u \incident e}\xb_u$. Here $\xb \in C^0(G, \Fs)$ is a 0-cochain and $\xb_v \in \Fs(v)$ is the vector of $\xb$ at the node stalk $\Fs(v)$. 
\end{definition}

The co-boundary map $\delta$ allows us to construct the \emph{sheaf Laplacian operator} over a sheaf.

\begin{definition}
The sheaf Laplacian of a sheaf is a map $L_\Fs : C^0(G, \Fs) \to C^0(G, \Fs)$ defined as $L_\Fs = \delta^\top \delta$.
\end{definition}

The sheaf Laplacian is a symmetric positive semi-definite (by construction) block matrix. The diagonal blocks are $L_{\Fs_{v,v}} = \sum_{v \incident e} \Fs^\top_{v \incident e}\Fs_{v \incident e}$, while the off-diagonal blocks are $L_{\Fs_{v,u}}= -\Fs^\top_{v \incident e}\Fs_{u \incident e}$.

\begin{definition}
The \emph{normalised sheaf Laplacian} $\Delta_\Fs$ is defined as $\Delta_\Fs = D^{-\frac{1}{2}}L_\Fs D^{-\frac{1}{2}}$ where $D$ is the block-diagonal of $L_\Fs$.
\end{definition}

Although stalk dimensions are arbitrary, we work with node and edge stalks which are all $d$-dimensional for simplicity. This means that each restriction map is $d\times d$, and therefore so is each block in the sheaf Laplacian. With $n$ we denote the number of nodes in the underlying graph $G$, which results in our sheaf Laplacian having dimensions $nd\times nd$. 

If we construct a trivial sheaf where each stalk is isomorphic to $\mathbb{R}$ and the restriction maps are identity maps, then we recover the well-known $n\times n$ graph Laplacian from the sheaf Laplacian. This effectively means that the sheaf Laplacian generalises the graph Laplacian by considering a non-trivial sheaf on $G$.





\begin{definition}
The {\em orthogonal (Lie) group} of dimension $d$, denoted $O(d)$, is the group of $d \times d$ orthogonal matrices together with matrix multiplication.
\end{definition}

If we constrain the restriction maps in the sheaf to belong to the orthogonal group (i.e., $\Fs_{v \incident e} \in O(d)$), the sheaf becomes a \emph{discrete $O(d)$-bundle} and can be thought of as a discretised version of a tangent  bundle on a manifold. The sheaf Laplacian of the $O(d)$-bundle is equivalent to a {\em connection Laplacian} used by \citet{singer2012vector}. The orthogonal restriction maps describe how vectors are rotated when transported between stalks, in a way analogous to the transportation of tangent vectors on a manifold. 

Orthogonal restriction maps are advantageous because orthogonal matrices have fewer free parameters, making them more efficient to work with. The Lie group $O(d)$ has a $d(d - 1)/2$-dimensional manifold structure (compared to the $d^2$-dimensional general linear group describing all invertible matrices). In $d=2$, for instance, $2\times 2$ rotation matrices have only one free parameter (the rotation angle). 
%

\subsection{Neural Sheaf Diffusion}
We now discuss the existing sheaf-based machine learning models and their theoretical properties. Consider a graph $G = (V, E)$ where each node $v \in V$ has a $d$-dimensional feature vector $\xb_v \in \Fs(v)$. We construct an $nd$-dimensional vector $\xb \in C^0(G, \Fs)$ by column-stacking the individual vectors $\xb_v$. Allowing for $f$ feature channels, we produce the feature matrix $\Xb \in \R^{(nd) \times f}$. The columns of $\Xb$ are vectors in $C^0(G, \Fs)$, one for each of the $f$ channels.

\emph{Sheaf diffusion} is a process on $(G,\Fs)$ governed by the following differential equation:
\begin{talign}
\label{eq:sheaf-diffusion}
\Xb(0) = \Xb\text{,    } \quad  \dot{\Xb}(t) = - \Delta_\Fs \Xb(t),
\end{talign}
which is discretised via the explicit Euler scheme with unit step-size:
\begin{talign*}
\Xb(t+1) = \Xb(t) - \Delta_\Fs\Xb(t) = \left(\Ib_{nd} - \Delta_\Fs \right) \Xb(t)\end{talign*}

The model used by 
\citet{bodnar2022neural} for experimental validation was of the form 
\begin{equation}
    \dot{\Xb} = - \sigma \left(\Delta_{\Fs(t)} \left(\mathbf{I}_n \otimes \Wb_1 \right) \Xb(t) \Wb_2 \right)
\label{eq:continous-sheaf}
\end{equation}
%
%
%
where 
$\Wb_1$ and $\Wb_2$ are weight matrices, the restriction maps defining $\Delta_{\Fs(t)}$  are computed by a learnable parametric matrix-valued function 
$\Fs_{v \incident e:=(v,u)}= \boldsymbol{\Phi}(\xb_v, \xb_u)$, on which additional constraints (e.g., diagonal or orthogonal structure) can be imposed. 
Equation (\ref{eq:continous-sheaf}) was 
discretised as 
\begin{equation}
    \Xb_{t+1} = \Xb_t - \sigma \left(\Delta_{\Fs(t)} \left(\mathbf{I}_n \otimes \Wb^t_1 \right) \Xb_t \Wb^t_2 \right)
    \label{eq:discrete-sheaf}
\end{equation}

It is important to note that the sheaf $\Fs(t)$ and the weights $\Wb^t_1,\Wb^t_2$ in 
equation (\ref{eq:discrete-sheaf}) 
are time-dependent, 
meaning that the underlying ``geometry'' evolves over time.

\section{Connection Sheaf Laplacians}
\label{sec:new-technique}
The 
sheaf Laplacian $\Delta_{\Fs(t)}$ arises from the sheaf $\Fs(t)$ built upon the graph $G$, which in turn is determined by constructing the individual restriction maps $\Fs_{v \incident e}$. 
Instead of learning a parametric function 
$\Fs_{v \incident e:=(v,u)} = \boldsymbol{\Phi}(\xb_v, \xb_u)$ as done by \citet{bodnar2022neural}, we compute the restriction maps in a non-parametric manner at pre-processing time. In doing so, we avoid learning the maps by backpropagation. In particular the restriction maps we compute are orthogonal. We work with this class because it was shown to be more efficient when using the same stalk width as compared to other models in \citet{bodnar2022neural}, and due to the geometric analogy to parallel transport on manifolds. 

\subsection{Local PCA \& Alignment for Point Clouds}
We adapt a procedure to learn orthogonal transformations 
on point clouds, 
presented by \citet{singer2012vector}. Their construction relies on the so-called ``manifold assumption'', positing that even though data lives in a high-dimensional space $\R^p$, the correlation between dimensions suggests that in reality, the data points lie on a $d$-dimensional Riemannian manifold 
$\Ms^d$ embedded in $\R^p$ (with significantly lower dimension, $d\ll p$). 

Assume the manifold 
$\Ms^d$ is sampled at points $\{ \mathbf{x}_1, \dots, \mathbf{x}_n \} \subset \mathbb{R}^p$. 
At every point $\mathbf{x}_i$, $\Ms^d$ has a {\em tangent space} $T_{\mathbf{x}_i}\Ms$ (which is analogous to our $\mathcal{F}(v)$) that intuitively contains all the vectors at $\mathbf{x}_i$ that are tangent to the manifold. 
A mechanism allowing to transport vectors between two $T_{\mathbf{x}_i}\Ms$ and $T_{\mathbf{x}_j}\Ms$ at nearby points is a {\em connection} (or {\em parallel transport}, which would correspond to our transport maps $\Fs^\top_{v \incident e}\Fs_{u \incident e}$ between $\mathcal{F}(u)$ and $\mathcal{F}(v)$).

Computing 
a connection on the discretised manifold is a two step procedure. First, orthonormal bases of the tangent spaces for each data point are constructed via local PCA. Next, the tangent spaces are optimally aligned via orthogonal transformations, which can be thought of as mappings from one tangent space to a neighbouring one.
%
\citet{singer2012vector} computed a $\sqrt{\epsilon_{PCA}}$-neighbourhood ball of points for each point $\mathbf{x}_i$ denoted $\Ns_{\mathbf{x}_i, \epsilon_{PCA}}$. This forms a set of neighbouring points $\mathbf{x}_{i_1}, \dots, \mathbf{x}_{i_{N_i}}$. Then the $p \times N_i$ matrix $\hat{\Xb}_i = [\mathbf{x}_{i_1} - \mathbf{x}_i, \ldots, \mathbf{x}_{i_{N_i}} - \mathbf{x}_i]$ is obtained, which centres all of the neighbours at $\mathbf{x}_i$. Next, an $N_i \times N_i$ weighting matrix $\Db_i$ is constructed, giving more importance to neighbours closer to $\mathbf{x}_i$. This allows us to compute the $p \times N_i$ matrix $\Bb_i = \hat{\Xb}_i \Db_i$. Then Singular Value Decomposition (SVD) is used on $\Bb_i$ such that $\Bb_i = \Ub_i \boldsymbol{\Sigma}_i \Vb_i^\top$. Assuming that the singular values are in decreasing order, the first $d$ left singular vectors are kept (the first $d$ vectors of $\mathbf{U}_i$), forming the matrix $\Ob_i$. Note that the columns of $\Ob_i$ are orthonormal by construction and they form a $d$-dimensional subspace of $\R^p$. This basis constitutes our approximation to the basis of the tangent space~$T_{\mathbf{x}_i}\Ms$.

 To compute the orthogonal matrix $\Ob_{ij}$, which represents our orthogonal transformation from $T_{\mathbf{x}_i}\Ms$ to $T_{\mathbf{x}_j}\Ms$, it is sufficient to first of all compute the SVD of $\Ob^\top_i \Ob_j = \Ub \boldsymbol{\Sigma} \Vb^\top$ and then $\Ob_{ij} = \Ub \Vb^\top$. $\Ob_{ij}$ is the orthogonal transformation which optimally aligns the tangent spaces $T_{\mathbf{x}_i}\Ms$ and $T_{\mathbf{x}_j}\Ms$ based on their bases $\Ob_i$ and $\Ob_j$. Whenever $\mathbf{x}_i$ and $\mathbf{x}_j$ are ``nearby'', \citet{singer2012vector} show that $\Ob_{ij}$ is an approximation to the parallel transport operator.

\subsection{Local PCA \& Alignment for Graphs}
The technique has many valuable theoretical properties, but was originally designed for point clouds. In our case, we also wish to leverage the valuable edge information at our disposal. To do this, instead of computing the neighbourhood $\Ns_{\mathbf{x}_i, \epsilon_{PCA}}$, we take the 1-hop neighbourhood $\Ns_{\mathbf{x}_i}^1$ of $\mathbf{x}_i$. A problem is encountered when computing the weighting matrix $\Db_i$, which gives different weightings dependent on the distance to the centroid of the neighbourhood. We make the assumption that $\Db_i$ is an identity matrix, giving the same weighting to each node in the neighbourhood, as they are all at a 1-hop distance from the reference feature vector. This means that in our approach $\Bb_i = \hat{\Xb}_i \Db_i = \hat{\Xb}_i$. 

Following this modification, the technique matches the procedure proposed by \citet{singer2012vector}. We compute the SVD of $\Bb_i$ to extract $\Ob_i$ from the left singular vectors. We finally compute the orthogonal transport maps $\Ob_{ij}$ from the SVD of $\Ob^\top_i \Ob_j$. This gives a modified version of the alignment procedure, that is now graph-aware. To the best of our knowledge, this a novel technique to operate over graphs. A diagram of the newly proposed approach is displayed in Figure \ref{fig:local-pca}.

Estimating $d$ is non-trivial, that is, the dimension of the tangent space (in our case, the stalks). In fact, we are assuming that every neighbourhood is larger than $d$ or else $\Bb_i$ would have less than $d$ singular vectors, and our construction would be ill-defined. This is clearly not always the case for all $d$. While \citet{singer2012vector} proposed to estimate $d$ directly from the data, we leave $d$ as a tunable hyper-parameter. 

To solve the problem for nodes which have less than $d$ neighbours, we take the closest neighbours in terms of the Euclidean distance which are not in the 1-hop neighbourhood. In other words, when there are less than $d$ neighbours, we pick the remaining neighbours following the original procedure by \citet{singer2012vector}.  We note that one could try to consider an $n$-hop neighbourhood instead, in a similar fashion to $\epsilon_{PCA}$ in the original technique. Still, this comes with a larger computational overhead and complications related to the weightings. Furthermore, if a graph has a disconnected node, this would still be an issue. In practice, $d$ is kept small such that most nodes have at least $d$ edge-neighbours.

Algorithm \ref{alg:graph-new} shows the pseudo-code for our technique. In principle, the LocalNeighbourhood function selects the neighbours based on the 1-hop neighbourhood. If the number of these neighbours is less than the stalk dimension, we pick the closest neighbours based on the Euclidean distance, which are not in the 1-hop neighbourhood. Assuming unit cost for SVD, the run-time increases linearly with the number of data-points. Also, given that the approach here described is performed at pre-processing time, we are able to compute the sheaf Laplacian in a deterministic way in constant time during training. This removes the overhead required whilst backpropagating through the sheaf Laplacian to learn the parametric function $\mathbf{\Phi}$. It also helps counter issues related to overfitting, especially when the dimension of the stalks increases as we are removing the additional parameters which come with $\mathbf{\Phi}$, reducing model complexity.

\begin{algorithm}[tb]
   \caption{Local PCA \& Alignment for Graphs}
   \label{alg:graph-new}
\begin{algorithmic}
   \STATE {\bfseries Input:} feature matrix $\Xb$, EdgeIndex, stalk dimension $d$
   
   \STATE // Graph Local PCA
   \FOR{$i = 0$ {\bfseries to} len($\Xb)$}
    \STATE // 1-hop neighbourhood and closest vectors 
    \STATE // (Euclidean distance) if needed, centred at $x_i$
    \STATE $\hat{\Xb}_i =$ LocalNeighbourhood($\Xb$, EdgeIndex, i)
    \STATE $\Ub_i, \boldsymbol{\Sigma}_i, \Vb^\top_i =$ SVD($\hat{\Xb}_i)$
    \STATE // Choose first $d$ left singular vectors
    \STATE $\Ob_i = \Ub_i[:,:d]$
   \ENDFOR
   
   \STATE // Alignment
   \FOR{i, j {\bfseries in} EdgeIndex}
    \STATE $\Ub, \boldsymbol{\Sigma}, \Vb^\top =$ SVD$(\Ob_{i}^\top \Ob_{j})$
    \STATE $\Ob_{i j} = \Ub \Vb^\top$
   \ENDFOR
\end{algorithmic}
\end{algorithm}

\section{Evaluation}
\label{sec:eval}
\newcommand{\first}[1]{\mathbf{\textcolor{red}{#1}}}
\newcommand{\second}[1]{\mathbf{\textcolor{blue}{#1}}}
\newcommand{\third}[1]{\mathbf{\textcolor{violet}{#1}}}

\begin{table*}[t]
    \centering
    \caption{Accuracy $\pm$ variance for various node classification datasets and models. The datasets are sorted by increasing order of homophily. Our technique is denoted Conn-NSD, while the other Sheaf Diffusion models are Diag-NSD, O(d)-NSD and Gen-NSD. The top three models are coloured by \textbf{\textcolor{red}{First}}, \textbf{\textcolor{blue}{Second}} and \textbf{\textcolor{violet}{Third}}, respectively. The first section includes sheaf-based models, while the second includes other GNN models.
    }
    
    \resizebox{\textwidth}{!}{%
    \begin{tabular}{l ccccccccc}
    \toprule 
         &
         \textbf{Texas} &  
         \textbf{Wisconsin} & 
         \textbf{Film} &
         \textbf{Squirrel} &
         \textbf{Chameleon} &
         \textbf{Cornell} &
         \textbf{Citeseer} & 
         \textbf{Pubmed} & 
         \textbf{Cora} \\
         
         Homophily level &
         \textbf{0.11} &
         \textbf{0.21} & 
         \textbf{0.22} & 
         \textbf{0.22} & 
         \textbf{0.23} &
         \textbf{0.30} &
         \textbf{0.74} &
         \textbf{0.80} &
         \textbf{0.81} \\ 
         
         \#Nodes &
         183 &
         251 & 
         7,600 &
         5,201 & 
         2,277 &
         183 &
         3,327 &
         18,717 &
         2,708 \\
         
         \#Edges &
         295 &
         466 & 
         26,752 & 
         198,493 & 
         31,421 &
         280 &
         4,676 &
         44,327 &
         5,278 \\
         
         \#Classes &
         5 &
         5 & 
         5 &
         5 &
         5 & 
         5 &
         7 &
         3 &
         6 
         
         \\ \midrule
         \textbf{Conn-NSD (ours)} &
         $\first{86.16} {\scriptstyle \pm 2.24}$ &
         $\third{88.73} {\scriptstyle \pm 4.47}$ &
         $\first{37.91} {\scriptstyle \pm 1.28}$ & 
         $45.19 {\scriptstyle \pm 1.57}$ & 
         $65.21 {\scriptstyle \pm 2.04}$ &
         $\second{85.95} {\scriptstyle \pm 7.72}$ & 
         $75.61 {\scriptstyle \pm 1.93}$ &
         $89.28 {\scriptstyle \pm 0.38}$ &
         $83.74 {\scriptstyle \pm 2.19}$\\

        RandEdge-NSD &
         $84.05 {\scriptstyle \pm 5.33}$ &
         $85.69 {\scriptstyle \pm 4.02}$ &
         $37.40 {\scriptstyle \pm 1.18}$ & 
         $33.89 {\scriptstyle \pm 1.56}$ & 
         $47.72 {\scriptstyle \pm 1.60}$ &
         $84.59 {\scriptstyle \pm 7.65}$ & 
         $72.49 {\scriptstyle \pm 1.91}$ &
         $87.74 {\scriptstyle \pm 0.50}$ &
         $74.00 {\scriptstyle \pm 1.99}$\\
         
        RandNode-NSD &
         $82.97 {\scriptstyle \pm 7.55}$ &
         $86.47 {\scriptstyle \pm 4.51}$ &
         $37.54 {\scriptstyle \pm 1.32}$ & 
         $34.00 {\scriptstyle \pm 1.43}$ & 
         $50.68 {\scriptstyle \pm 2.48}$ &
         $83.78 {\scriptstyle \pm 7.81}$ & 
         $73.89 {\scriptstyle \pm 1.94}$ &
         $89.13 {\scriptstyle \pm 0.59}$ &
         $80.90 {\scriptstyle \pm 1.51}$\\
         
         Diag-NSD &
         $\third{85.67} {\scriptstyle \pm 6.95}$ &
         $88.63 {\scriptstyle \pm 2.75}$ &
         $37.79 {\scriptstyle \pm 1.01}$ & 
         $\third{54.78} {\scriptstyle \pm 1.81}$ & 
         $\second{68.68} {\scriptstyle \pm 1.73}$ &
         $\first{86.49} {\scriptstyle \pm 7.35}$ & 
         $\third{77.14} {\scriptstyle \pm 1.85}$ &
         $89.42 {\scriptstyle \pm 0.43}$ &
         $87.14 {\scriptstyle \pm 1.06}$\\
         
         $O(d)$-NSD &
         $\second{85.95} {\scriptstyle \pm 5.51}$ &
         $\first{89.41} {\scriptstyle \pm 4.74}$ &
         $\second{37.81} {\scriptstyle \pm 1.15}$ & 
         $\first{56.34} {\scriptstyle \pm 1.32}$  & 
         $\third{68.04} {\scriptstyle \pm 1.58}$ &
         $84.86 {\scriptstyle \pm 4.71}$ & 
         $76.70 {\scriptstyle \pm 1.57}$ &
         $\third{89.49} {\scriptstyle \pm 0.40}$ &
         $86.90 {\scriptstyle \pm 1.13}$ \\
         
         Gen-NSD &
         $82.97 {\scriptstyle \pm 5.13}$ &
         $\second{89.21} {\scriptstyle \pm 3.84}$ &
         $\third{37.80} {\scriptstyle \pm 1.22}$ & 
         $53.17 {\scriptstyle \pm 1.31}$ & 
         $67.93 {\scriptstyle \pm 1.58}$ &
         $\third{85.68} {\scriptstyle \pm 6.51}$ & 
         $76.32 {\scriptstyle \pm 1.65}$ &
         $89.33 {\scriptstyle \pm 0.35}$ &
         $87.30 {\scriptstyle \pm 1.15}$ \\
        
        \midrule
         GGCN &
         $84.86 {\scriptstyle \pm 4.55}$ &
         $86.86 {\scriptstyle \pm 3.29}$ &
         $37.54 {\scriptstyle \pm 1.56}$ &
         $\second{55.17} {\scriptstyle \pm 1.58}$ & 
         $\first{71.14} {\scriptstyle \pm 1.84}$ &
         $\third{85.68} {\scriptstyle \pm 6.63}$ &
         $\third{77.14} {\scriptstyle \pm 1.45}$ &
         $89.15 {\scriptstyle \pm 0.37}$ &
         $\second{87.95} {\scriptstyle \pm 1.05}$ \\
         
         H2GCN &
         $84.86 {\scriptstyle \pm 7.23}$ &
         $87.65 {\scriptstyle \pm 4.98}$ &
         $35.70 {\scriptstyle \pm 1.00}$ &
         $36.48 {\scriptstyle \pm 1.86}$ & 
         $60.11 {\scriptstyle \pm 2.15}$ &
         $82.70 {\scriptstyle \pm 5.28}$ &
         $77.11 {\scriptstyle \pm 1.57}$ &
         $\third{89.49} {\scriptstyle \pm 0.38}$ &
         $\third{87.87} {\scriptstyle \pm 1.20}$ \\

         GPRGNN &
         $78.38 {\scriptstyle \pm 4.36}$ &
         $82.94 {\scriptstyle \pm 4.21}$ &
         $34.63 {\scriptstyle \pm 1.22}$ & 
         $31.61 {\scriptstyle \pm 1.24}$ &
         $46.58 {\scriptstyle \pm 1.71}$ &
         $80.27 {\scriptstyle \pm 8.11}$ &
         $77.13 {\scriptstyle \pm 1.67}$ &
         $87.54 {\scriptstyle \pm 0.38}$ &
         $\second{87.95} {\scriptstyle \pm 1.18}$ \\ 
         
         FAGCN &
         $82.43 {\scriptstyle \pm 6.89}$ &
         $82.94 {\scriptstyle \pm 7.95}$ &
         $34.87 {\scriptstyle \pm 1.25}$ &
         $42.59 {\scriptstyle \pm 0.79}$ &
         $55.22 {\scriptstyle \pm 3.19}$ & 
         $79.19 {\scriptstyle \pm 9.79}$ &
         N/A & 
         N/A & 
         N/A \\
         
         MixHop &
         $77.84 {\scriptstyle \pm 7.73}$ &
         $75.88 {\scriptstyle \pm 4.90}$ &
         $32.22 {\scriptstyle \pm 2.34}$ & 
         $43.80 {\scriptstyle \pm 1.48}$ &
         $60.50 {\scriptstyle \pm 2.53}$ &
         $73.51 {\scriptstyle \pm 6.34}$ &
         $76.26 {\scriptstyle \pm 1.33}$ &
         $85.31 {\scriptstyle \pm 0.61}$ &
         $87.61 {\scriptstyle \pm 0.85}$ \\

         GCNII &
         $77.57 {\scriptstyle \pm 3.83}$ &
         $80.39 {\scriptstyle \pm 3.40}$  &
         $37.44 {\scriptstyle \pm 1.30}$ &
         $38.47 {\scriptstyle \pm 1.58}$ &
         $63.86 {\scriptstyle \pm 3.04}$ &
         $77.86 {\scriptstyle \pm 3.79}$ &
         $\second{77.33} {\scriptstyle \pm 1.48}$ &
         $\first{90.15} {\scriptstyle \pm 0.43}$ &
         $\first{88.37} {\scriptstyle \pm 1.25}$ \\
         
         Geom-GCN &
         $66.76 {\scriptstyle \pm 2.72}$ &
         $64.51 {\scriptstyle \pm 3.66}$ &
         $31.59 {\scriptstyle \pm 1.15}$ & 
         $38.15 {\scriptstyle \pm 0.92}$ & 
         $60.00 {\scriptstyle \pm 2.81}$ &
         $60.54 {\scriptstyle \pm 3.67}$ &
         $\first{78.02} {\scriptstyle \pm 1.15}$ &
         $\second{89.95} {\scriptstyle \pm 0.47}$ &  
         $85.35 {\scriptstyle \pm 1.57}$ \\ 
         
         PairNorm &
         $60.27 {\scriptstyle \pm 4.34}$ &
         $48.43 {\scriptstyle \pm 6.14}$ &
         $27.40 {\scriptstyle \pm 1.24}$ & 
         $50.44 {\scriptstyle \pm 2.04}$ & 
         $62.74 {\scriptstyle \pm 2.82}$ &
         $58.92 {\scriptstyle \pm 3.15}$ &
         $73.59 {\scriptstyle \pm 1.47}$ &
         $87.53 {\scriptstyle \pm 0.44}$ &
         $85.79 {\scriptstyle \pm 1.01}$ \\ 
         
         GraphSAGE &
         $82.43 {\scriptstyle \pm 6.14}$ &
         $81.18 {\scriptstyle \pm 5.56}$ &
         $34.23 {\scriptstyle \pm 0.99}$ & 
         $41.61 {\scriptstyle \pm 0.74}$ & 
         $58.73 {\scriptstyle \pm 1.68}$ &
         $75.95 {\scriptstyle \pm 5.01}$ &
         $76.04 {\scriptstyle \pm 1.30}$ &
         $88.45 {\scriptstyle \pm 0.50}$ &
         $86.90 {\scriptstyle \pm 1.04}$\\
         
         GCN &
         $55.14 {\scriptstyle \pm 5.16}$ &
         $51.76 {\scriptstyle \pm 3.06}$ &
         $27.32 {\scriptstyle \pm 1.10}$ & 
         $53.43 {\scriptstyle \pm 2.01}$ &
         $64.82 {\scriptstyle \pm 2.24}$ &
         $60.54 {\scriptstyle \pm 5.30}$ &
         $76.50 {\scriptstyle \pm 1.36}$ &
         $88.42 {\scriptstyle \pm 0.50}$ &
         $86.98 {\scriptstyle \pm 1.27}$ \\ 
         
         GAT &
         $52.16 {\scriptstyle \pm 6.63}$ &
         $49.41 {\scriptstyle \pm 4.09}$ &
         $27.44 {\scriptstyle \pm 0.89}$ & 
         $40.72 {\scriptstyle \pm 1.55}$ &
         $60.26 {\scriptstyle \pm 2.50}$ &
         $61.89 {\scriptstyle \pm 5.05}$ &
         $76.55 {\scriptstyle \pm 1.23}$ &
         $87.30 {\scriptstyle \pm 1.10}$ & 
         $86.33 {\scriptstyle \pm 0.48}$ \\ 
         
         MLP &
         $80.81 {\scriptstyle \pm 4.75}$ &
         $85.29 {\scriptstyle \pm 3.31}$ &
         $36.53 {\scriptstyle \pm 0.70}$ & 
         $28.77 {\scriptstyle \pm 1.56}$ & 
         $46.21 {\scriptstyle \pm 2.99}$ &
         $81.89 {\scriptstyle \pm 6.40}$ &
         $74.02 {\scriptstyle \pm 1.90}$ &
         $75.69 {\scriptstyle \pm 2.00}$ & 
         $87.16 {\scriptstyle \pm 0.37}$ \\ 

         \bottomrule
         
    \end{tabular}
    }
    \label{tab:main-results}
\end{table*}

We evaluate our model on several datasets, and compare its performance to a variety of models recorded in the literature, as well as to some especially designed baselines. For consistency, we use the same datasets as the ones discussed by \citet{bodnar2022neural}. These are real-world datasets which aim at evaluating heterophilic learning \cite{rozemberczki2021multi, pei2020geom}. They are ordered based on their homophily coefficient $0 \leq h \leq 1$, which is higher for more homophilic datasets. Effectively, $h$ is the fraction of edges which connect nodes of the same class label. The results are collected over $10$ fixed splits, where 48\%, 32\%, and 20\% of nodes per class are used for training, validation, and testing, respectively. The reported results are chosen from the highest validation score.

Table \ref{tab:main-results} contains accuracy results for a wide range of models, along with ours, Conn-NSD, for node classification tasks. An important baseline is the Multi-Layer Perceptron (MLP), whose result we report in the last row of Table \ref{tab:main-results}. The MLP has access only to the node features and it provides an idea of how much useful information GNNs can extract from the graph structure. The GNN models in Table \ref{tab:main-results} can be classiffied in 3 main categories:
\begin{enumerate}
    \item Classical: GCN \cite{kipf2016semi}, GAT \cite{velickovic2017graph}, GraphSAGE \cite{hamilton2017inductive},
    \item Models for heterophilic settings: GGCN \cite{yan2021two}, Geom-GCN \cite{pei2020geom}, H2GCN \cite{zhu2020generalizing}, GPRGNN \cite{chien2020joint}, FAGCN \cite{bo2021beyond}, MixHop \cite{abu2019mixhop},
    \item Models which address over-smoothing: GCNII \cite{chen2020simple}, PairNorm \cite{zhao2019pairnorm},
\end{enumerate}
Additionally, we also include the results presented by \citet{bodnar2022neural} using sheaf diffusion models, and the two random baselines: RandEdge-NSD and RandNode-NSD. RandEdge-NSD generates the sheaf by sampling a Haar-random matrix \citep{meckes2019random} for each edge. RandNode-NSD instead generates the sheaf by sampling a Haar-random matrix for each node $\Ob_i$ and then by computing the transport maps $\Ob_{ij}$ from $\Ob_i$ and $\Ob_j$. These last two baselines help us determine how our sheaf structure performs against a randomly sampled one.  

As we can see from the results, sheaf diffusion models tend to perform best for the heterophilic datasets such as Texas, Wisconsin, and Film. On the other hand, their relative performance drops as homophily increases. This is expected since, for example, classical models such as GCN and GAT exploit homophily by construction, whereas sheaf diffusion models are more general, adaptable, and versatile, but at the same time lose the inductive bias provided by classical models for homophilic data. 

Conn-NSD, alongside the other original discrete sheaf diffusion methods, consistently beats the random orthogonal sheaf baselines, which shows that our model incorporates meaningful geometric structure. The proposed Conn-NSD model achieves excellent results on the Texas and Film datasets, outperforming Diag-NSD, O(d)-NSD, and Gen-NSD, using fewer learnable parameters. Furthermore, Conn-NSD also obtains competitive results for Wisconsin, Cornell and Pubmed and remains close-behind on Citeseer and Cora. 

It is only in the case of the Squirrel dataset, and to a lesser extent Chameleon, that Conn-NSD is not able to perform as well as the models discussed by \citet{bodnar2022neural}. The Squirrel dataset contains a large amount of nodes and a substantially greater number of edges than all the other datasets. Importantly, the underlying MLP used for classification scores poorly. It may be that the extra flexibility provided by learning the sheaf is specially beneficial in cases in which the underlying MLP achieves low accuracy.  Nevertheless, Conn-NSD still convincingly outperforms the random baselines, especially on these last two datasets.

\begin{table*}[t]
    \caption{Mean seconds per epoch for each of the datasets. The proposed model achieves faster inference times because it does not need to learn and build a Laplacian at each layer.}
    \centering
    \resizebox{0.8\textwidth}{!}{%
    \begin{tabular}{l ccccccccc}
    \toprule 
         &
         \textbf{Texas} &  
         \textbf{Wisconsin} & 
         \textbf{Film} &
         \textbf{Squirrel} &
         \textbf{Chameleon} &
         \textbf{Cornell} &
         \textbf{Citeseer} & 
         \textbf{Pubmed} & 
         \textbf{Cora} \\
         
         \#Nodes &
         183 &
         251 & 
         7,600 &
         5,201 & 
         2,277 &
         183 &
         3,327 &
         18,717 &
         2,708 \\
         
         \#Edges &
         295 &
         466 & 
         26,752 & 
         198,493 & 
         31,421 &
         280 &
         4,676 &
         44,327 &
         5,278 \\
         
         
         \midrule
         \textbf{Conn-NSD (ours)} &
         $0.010$ &
         $0.013$ &
         $0.017$ & 
         $0.310$ & 
         $0.169$ &
         $0.013$ & 
         $0.011$ &
         $0.147$ &
         $0.015$\\

         $O(d)$-NSD &
         $0.017$ &
         $0.018$ &
         $0.022$ & 
         $0.572$  & 
         $0.296$ &
         $0.019$ & 
         $0.017$ &
         $0.263$ &
         $0.022$ \\

         \bottomrule
         
    \end{tabular}
    }
    \label{tab:runtime_results}
\end{table*}
Overall, Conn-NSD performs comparably well to learning the sheaf via gradient-based approaches in most cases. It also seems most well-suited on graphs with a very low amount of nodes. This may be explained by the fact that Conn-NSD aims to mitigate overfitting, acting as a form of regularisation which allows for faster training and fewer parameters. 

\paragraph{Runtime performance} Finally, we measure the speedup achieved by moving the computation of the sheaf Laplacian at pre-processing time. Table \ref{tab:runtime_results} displays the mean wall-clock time for an epoch measured in seconds, obtained with a NVIDIA TITAN X GPU and an Intel(R) Core(TM) i7-6700 CPU @ 3.40GHz. Conn-NSD achieves significantly faster inference times when compared to its direct counter-part $O(d)$-NSD from \citet{bodnar2022neural}. The larger datasets see the most benefit, with Squirrel showing a $45.8\%$ speed up. 

\section{Conclusion}
We proposed and evaluated a novel technique to compute the sheaf Laplacian of a graph deterministically, obtaining promising results. This was done by leveraging existing differential geometry work that constructs orthogonal maps that optimally align tangent spaces between points, relying on the manifold assumption. We crucially adapted this intuition to be graph-aware, leveraging the valuable edge connection information in the graph structure. 

We showed that this technique achieves competitive empirical results and it is able to beat or match the performance of the original models by \citet{bodnar2022neural} on most datasets, as well as to consistently outperform the random sheaf baselines. This suggests that in some cases it may not be necessary to learn the sheaf through a parametric function, but instead the sheaf can be computed as a pre-processing step. This work may be regarded as a regularisation technique for SNNs, which also reduces the training time as it removes the need to backpropagate through the sheaf.

We believe we have uncovered an exciting research direction which aims to find a way to compute sheaves non-parametrically with an objective that is independent of the downstream task. Furthermore, we are excited by the prospect of further research tying intuition stemming from the fields of algebraic topology and differential geometry to machine learning. We believe that this work forms a promising first step in this direction.




\nocite{langley00}

\bibliography{bib}
\bibliographystyle{icml2022}

\newpage
\appendix
\appendix
\onecolumn

\end{document}